# Geospatial semantics: beyond ontologies, towards an enactive approach

Pasquale Di Donato

LABSITA – Sapienza University of Rome,
Piazza Borghese 9, 00186 Rome, Italy.
pasquale.didonato@uniroma1.it

**Abstract**. Current approaches to semantics in the geospatial domain are mainly based on ontologies, but ontologies, since continue to build entirely on the symbolic methodology, suffers from the classical problems, e.g. the symbol grounding problem, affecting representational theories. We claim for an enactive approach to semantics, where meaning is considered to be an emergent feature arising context-dependently in action. Since representational theories are unable to deal with context, a new formalism is required toward a contextual theory of concepts. SCOP is considered a promising formalism in this sense and is briefly described.

## 1 Introduction

The current scene of Geographic Information (GI) is characterised by the provision of services, in a distributed information systems environment, that enable to integrate distributed information resources

Dealing with data integration basically implies addressing two main types of heterogeneity: data heterogeneity and semantic heterogeneity. Data heterogeneity refers to differences in data in terms of data type and data formats, while semantic heterogeneity applies to the meaning of the data (Hakimpour 2001); semantic heterogeneity may consists of naming heterogeneities -when different words/expressions are used for the same (semantically alike) concept- and conceptual heterogeneities -when different people and or disciplines have a different interpretation, conceptualisation of the same "thing"- (Bishr 1998).

The Open Geospatial Consortium and the ISO TC 211 provide specifications and standards supporting the deployment of geospatial web services. These specifications and standards address the interoperability issue at syntactic level, but are limited in terms of semantics and do not provide a consistent model for the semantics integration/composition of geospatial services (Einspanier 2003).

Coping with semantic interoperability is a challenging task, since it has more to do with how people perceive and give meaning to "things", rather then with integrating software components through standard interfaces (Harvey 1999).

Semantics deals with aspects of meaning as expressed in a language, either natural or technical such as a computer language, and is complementary to syntax which deals with the structure of signs (focusing on the form) used by the language itself. In the area of distributed data sources and services, semantic interoperability refers to the ability of systems to exchange data and functionalities in a meaningful way. Semantic heterogeneity occurs when there is no agreement of the meaning of the same data and/or service functionality.

Data creation happens in a context or in an application domain where concepts and semantics are clear to the data creator, either because they are explicitly formalised either they are naturally applied due to a yearly experience. But with distributed data resources this context is missed and unknown to the end user. This means that, in order to achieve semantic interoperability, semantics should be formally and explicitly represented (Kuhn 2005).

Current approaches to overcome semantic heterogeneities rely on the use of ontologies and reasoning engines for concepts matching among different ontologies. The main drawback is that ontologies, being forms of *a priori* agreements, are decontextualised and decontextualise experience; instead our assumption is that semantics reconciliation depends on contextual human sense-making. This claims for a new formalism for geospatial semantics.

The rest of this paper is organised as follows: section 2 deals with the semantics issue in the broader context of cognitive sciences; section 3 briefly summarises current approaches to semantics in distributed information systems; section 4 summarises current approaches to semantics in the GI arena; section 5 introduces the SCOP formalism

## 2 Semantics

There is a general agreement on considering that semantics deals with relationships between linguistic expressions and their meaning; but when it turns do define such relationships opinions highly diverge (Gärdenfors 2000) in a dispute which is mainly philosophical.

Dealing with formal semantics means opting for one of the two following paradigms, and the choice is mostly philosophical:

- Realistic semantics, which comes in two flavours:

  - Extensional: in extensional semantics terms of a language **L** are mapped onto a "world" **W**. The main aim is to determine truth conditions for sentences in **L** against **W**. Extensional semantics is rooted in Tarski's model theory for first order logic, where sentences from a language get their meaning via a correspondence to a model assumed to be a representation of the world: this meaning is independent of how people understand it;

  - Intentional: in intentional semantics the language **L** is mapped onto a set of *possible worlds*, and the aim continues to be that of providing truth conditions for sentences in **L**;

- Cognitive semantics: in cognitive semantics the meanings of sentences are "mental", and linguistic expressions are mapped onto cognitive structures. According to Gärdenfors (2000): (i) meaning is a conceptualisation in a cognitive model and is independent of truth; (ii) cognitive models are mainly perceptual; (iii) semantic elements are based on spatial and topological objects, and are not symbols; (iv) cognitive models are image schematic and not propositional; (v) semantics is primary to syntax; (vi) concepts show prototype effects.

In order to analyse the semantics issue in a broader context we need to shift our focus to developments in the field of cognitive sciences. A critical review of cognitive sciences evolution can be found, for example, in (Anderson 2003), (Froese 2007), (Steels 2007), (Dreyfus 2007), (Gärdenfors 1999), (Licata 2008).

Cognitive Sciences is an interdisciplinary field of investigation, with ideas coming from several disciplines such as philosophy, psychology, neurosciences, linguistics, computer science, anthropology, biology, and physics. The main aim of cognitive sciences is trying to answer questions such as "What is reason? How do we make sense of our experience? What is a conceptual system and how is organised? Do all people use the same conceptual system?" (Lakoff 1987, xi)

Two main different approaches try to answer these questions in different ways. The traditional approach claims that reasoning/cognition is essentially a form of information processing oriented to problem solving. It comes in two flavours: *cognitivism*, which sees the human brain as a deterministic machine manipulating symbols in an algorithmic way; *connectionism*, which uses a sub-symbolic representation and considers cognition as emergent from a network of atomic components (Artificial Neural Networks).

The new approach is characterised by the so-called embodied-embedded mind hypothesis and its variants and extensions, such as *situated cognition*, and *enactivism*. Two terms are borrowed from Lakoff (1987) to indicate the two approaches; *objectivism* for the traditional approach, and *experiential realism* for the new approach:

## 2.1 Objectivism

Objectivism assumes that reason and cognition consist of symbols manipulation, where symbols get meaning through a correspondence to the *real world* (or *possible worlds*) objectively defined and independent of any interaction with human beings: incidentally this means that cognition is substantially context-free. Since the approach involves computation, it is also known as *computationalism* (Licata 2008), (Gärdenfors 2000).

Objectivism is rooted in the logical positivism wave of analytical philosophy as formalised at the beginning of the 20$^{th}$ century by the Vienna Circle. The assumption is that scientific reasoning is based on observational data derived from experiments: new knowledge is acquired from data through logically valid inferences. Only hypothesis grounded in first-order logic with model-theoretic interpretations – or some equivalent formalism – have a scientific validity.

Objectivism is also *reductionist* since it assumes that a system can be totally analysed and defined in terms of its components, in a kind of *divide et impera* process. The main fallacy of the *reductionist* hypothesis is to give for granted the reversibility of the process, but the hypothesis is not necessarily *constructionist*:

> "… the more the elementary particle physicists tell us about nature of the fundamental laws, the less relevance they seem to have [...] to the rest of science. [...] The behaviour of large and complex aggregates of elementary particles [...] is not to be understood in terms of a simple extrapolation of the properties of few particles. [...] at each level of complexity entirely new properties appear ..."

(Anderson 1972, 393).

We may summarise the fundamental views of objectivism as follows (Lakoff 1987):

- the mind can be considered as a computer (a Turing machine);
- symbols get their meaning in relation to "things" in the real world, thus they are internal representations of an external objective reality independent of human being;
- thought is context-free and disembodied;
- categorisation is the way we make sense of experience and categories are defined via sharing necessary and sufficient memberships properties;
- category symbols are grounded (get their meaning) in categories existing in the world independent of human being.

Dreyfus, especially with his often cited book *What Computers Can't Do: A Critique of Artificial Reason*, has strongly criticised computationalism; his criticism is inspired by the Heideggerian criticism to the reductionist position, rooted in the Cartesian philosophy, of computationalism. According to Heidegger cognition is the result of our experience in *being-in-the-world*, and is grounded in our disposition to react in a flexible way as required by a specific context (Froese 2007).

Guided by this idea, Dreyfus (2007) claims that the representation of significance and relevance is the main problem of computationalism: assuming that a computer stores facts about the world, how can it manage to know which facts are relevant in any given situation?.

A version of the relevance problem is the well known *frame problem* (Dennet 1984), (McCarthy 1969), i.e. the problem for a computer, running a representation of the world, in managing world changes: which changes are relevant for the new situation? Which have to be retained, since relevant? How to determine what is relevant and what is not?

The frame problem may be considered as a manifestation of symptoms of the *symbol grounding problem* (Harnad 1990). Harnad questions the claim of computationalism that semantic interpretation of a formal symbolic system is intrinsic to the system itself; instead, he claims that meaningless symbols, manipulated on the basis of their shape, are grounded in anything but other meaningless symbols. Imagine we have a calculator and type *2+2=4*: it is undoubted that this makes sense, but it makes sense in our head and not in the calculator.

Furthermore, a symbolic system is vulnerable to the Searle's Chinese Room argument (Harnad 1994), discussed in (Searle 1980) as a criticism

to the computationalism position according to which an appropriately programmed computer is a mind and has cognitive states.

Another questioned point is related to the traditional view of categories: on the objectivist view a category is defined as a set of "things" that share certain properties, which are necessary and sufficient conditions for defining the category itself. Implications of this position are: (i) members of a category are all equivalent, there is not a better example; (ii) categories are independent of any peculiarity of people doing the categorisation; (iii) new categories are derived through the composition of existing categories on a set-theoretic base.

According to Rosch (1978) most categories do not have clear-cut boundaries and exhibit a prototype effect, i.e. some members are better examples of a category than others ("apple" is a better example of the category "fruit" than "fig"). Rosch refers to a perceived world, rather than a metaphysical one without a knower: an object is put in a category on the basis of a *similarity judgement* to the category prototype as perceived by a knower. This means that human capacities to perceive, to form mental images, to organise "things" play their role in categorisation; categories are culture-dependent conventions shared by a given group of people, and become accepted at global level through a communicative process (Licata 2008), (Gärdenfors 2000).

As a first alternative approach to computationalism, connectionism is based on the assumption that we need to simulate the brain structure and functioning in order to understand cognition. Connectionism claims that cognition is the result of the emergence of global states from a network of simple connected components: the focus is primarily on learning, rather than on problem solving. Several authors have debated if this is actually a new paradigm or rather a different approach to the implementation of classical systems. What differentiates connectionism from computationalism is basically the functional architecture of the computation (parallel vs. serial), and the nature of the representation (sub-symbolic vs. symbolic); for both connectionism and computationalism cognition is basically a form of information processing. Both approaches lack of embodiment and autonomy (Searle 1980).

## 2.2 Experiential realism

Different studies in anthropology, linguistics, psychology show results in conflict with the objectivist view of the mind; the evidence suggests a different view of human cognition, whose characteristics are briefly summarized:

- Mind is embodied, meaning that our cognitive system is determined by our body. Our thoughts, ideas, concepts, and other aspects of our mind are shaped by our body, by our perceptual system, by our activity and experience in the world (Lakoff 1987);
- Thought is imaginative, and employs metaphors, metonymies and image schemas. This imaginative capacity is embodied since metaphors, metonymies and images are often based on bodily experience (Lakoff 1987);
- Thought has gestalt properties, i.e. the way our brain operates is holistic, parallel, and analogue with self-organising tendencies, thus it is not atomistic (Lakoff 1987);
- Cognitive models are mainly image-schematic and not propositional. Metaphors and metonymies, which are considered exceptional features on the traditional view, play a foundational role since they are transformations of image-schemas (Gärdenfors 2000);
- Concepts show prototype effects (Gärdenfors 2000);
- Meaning is a conceptual structure in a cognitive system (Gärdenfors 2000).

Experiential realism claims that cognitive processes emerge from real-time, goal-directed interactions between agents and their environment.

Experiential realism and the embodied hypothesis root their basis in the Heideggerian philosophy. In Being and Time (Heidegger 1927) Heidegger claims that the world is experienced as a significant whole and cognition is grounded in our skilful disposition to respond in a flexible and appropriate way to the significance of the contextual situation.

The world is made up of possibilities for action that require appropriate responses. Things in the world are not experienced in terms of entities with functional characteristics; rather our experience when pressed into possibilities deals directly responding to a "what-for": thus, a hammer is "for" hammering and our action of hammering discovers the *readiness-to-hand* of the hammer itself. The *readiness-to-hand* is not a fixed functionality or characteristic encountered in a specific situation; rather it is experienced as a solicitation that requires a flexible response to the significance of a specific context (Dreyfus 2007). This is similar to what Gibson calls "affordances" (Gibson 1977). Affordance is what matter when we are confronting an environment; we experience entities of the environment that solicit us to act in a skilful way, rather than their physical features, which do not influence our action directly.

The notion of "image schemas" has been jointly introduced by Lakoff and Johnson as a fundamental pillar of experiential realism (Hampe 2005). Image schemas are recurring and dynamic patterns of out perceptual interactions with the world that give coherence to our experience; they are pre-conceptual structures directly meaningful since they are grounded in our bodily experience, and have an inherent spatial structure constructed from basic topological and geometrical structures, i.e. "*container*", "*source-path-goal*", "*link*" (Gärdenfors 2000). Metaphors and metonymies are considered as cognitive operations that transform image schemas from a source to a target mental space. Fauconnier and Turner call this process "conceptual blending" (Fauconnier 1994).

Since image schemas are conceptual structures, they pertain to a particular individual: the question is how individual mental spaces become shared conventions (Gärdenfors 2006).Recent studies demonstrate that conventions emerge out of a communicative process between agents (Steels 2006), (Loula 2005), (Puglisi 2008).

Puglisi (2008) shows how a common language may emerge as a result of a communicative dialogue. A simulation with an assembly of agents demonstrates that a simple negotiation scheme, based on game theory, may guarantee the emergence of a self-organised communication system capable of discriminating and categorising objects in the world with few linguistic categories: in the simulation individual agents are endowed with the ability to form perceptual categories, while interaction/communication among agents produces the emergence and alignment of the linguistic categories.

As Gärdenfors (2006) puts it, semantics, thus meaning, is a "meeting of mind" where a communicative interaction enables to reach a semantic equilibrium.

## 2.2 Enactivism

Several authors (Dreyfus 2007), (Di Paolo 2003, 2007), (Froese 2007), (De Jaegher 2007) argue that paradoxically this meeting of mind is what is actually missing in empirical and theoretical investigation of the embodied hypothesis, where the focus is rather on agent's individual cognitive mechanisms as a form of closed sensorimotor feedback loops.

Enactivism is an attempt to move embodied practices beyond their current focus (Froese 2007). Enactivism is a term used by Maturana, Varela, Thomson, and Rosch to name their theories and is closely related to experiential realism; it is not a radically new idea, rather it is a synthesis of different ideas and approaches (Maturana 1980). Enactivism is characterised by five main ideas (Di Paolo 2007):
- Autonomy: cognising organisms are autonomous by virtue of their self-generated identity. A system whose identity is specified by a designer cannot exhibit autonomy since it can only "obey" to rules imposed in the design. Autonomous agency emphasises the role of the cogniser in determining the rules of the "game" being played;
- Sense-making: in our *being-in-the-world* we try to preserve our self-generated identity through interactions and exchanges with the environment which are significant for us; we actively participate in the creation of meaning via our action, rather than passively receive stimulus from the environment and create internal "images" of it. De Jaegher (2007) further extends the notion of sense-making into the realm of social cognition, where the active coupling of an agent is with another agent. This is what Barsalou (2003) calls social-embodiment ;
- Emergence: the notions of autonomy and sense-making invoke emergence. Autonomy is not a property of something, but the result of a new identity that emerges out of dynamical processes. Emergence means the formation of a new property or process with its own autonomous identity out of the interaction of existing processes;
- Embodiment: for enactivism cognition is embodied action, temporally and spatially embedded. Reasoning, problem solving, and mental images manipulation depend on bodily structures
- Experience: experience is a skilful aspect of embodied activity. As we progress from beginners to experts our performance improves, but experience also changes.

The following table is a short synopsis of the different approaches to cognition:

|  | **Computationalism** | **Connectionism** | **Embodiment/ Enactivism** |
| --- | --- | --- | --- |
| **Metaphor for the mind** | Mind as computer (Turing machine) | Mind as parallel distributed network | Mind inseparable from experience and world |
| **Metaphor for cognition** | Rule-based manipulation of symbols | Emergence of global states in a network of simple components | Ongoing interaction with the world and with other agents |
| **The world** | Separate and objective. Re-presentable via symbols | Separate and objective Re-presentable via patterns on network activation | Engaged  Presentable through action |
| **Mind/body** | Separable | Separable | Inseparable |

Table 1: synopsis of different approaches to cognition

## 3 Dealing with semantics in information systems

In the domain of distributed information systems the role of semantics for the automatic/semi-automatic exploitation of distributed resources is particularly emphasised. This process requires semantic interoperability, i.e. the capability of an information system to understand the semantic of a user request against that of an information source and mediate among them (Sheth 1999).

But how can semantics be specified? Uschold (2003) proposes the following classification: (i) implicit semantics; (ii) informally expressed semantics; (iii) formally expressed semantics for human consumption; (iv) formally expressed semantics for machine processing.

In (Sheth 2005) the following classification is proposed:

- Implicit semantics: is the semantic implicit in patterns in data and not explicitly represented, i.e. co-occurrence of documents or terms in clusters, hyperlinked documents;
- Formal semantics in the form of ontologies: in order to be machine readable and processable, semantics need to be represented in some sort of formalism. Formal languages are based on the notion of Model and Model Theory: expressions in the language are interpreted in models assumed to be representations of the world or of possible worlds. Description logics (DLs) is the current dominant formalism: based on sets theory, it has the drawback of being not able to represent graded concept membership and uncertainty;
- Powerful (soft) semantics: implies the use of fuzzy or probabilistic mechanisms to overcome the rigid interpretations of set-based mechanisms, and enables to represent degrees of memberships and certainty.

In an environment of distributed, heterogeneous data sources and information systems, semantic interoperability refers to the ability of systems to exchange data and software functionalities in a meaningful way; semantic heterogeneity, i.e. naming and conceptual conflicts arises when there is no agreement on the meaning of the same data and/or software functionality.

Explicit and formal semantic is seen as a solution to the problem, and this has motivated several authors to apply formal ontologies (Guarino 1998). Current practices, therefore, rely on ontologies creation and automated resources annotation, coupled with appropriate computational approaches, such as reasoning and query processing, for concept matching among different ontologies and against user queries (Sheth 2005).

Notwithstanding ontologies are seen as a solution to semantic heterogeneity, the irony is that a clear understanding of ontology itself is far to be achieved (Agarwal 2005), and this understanding varies across disciplines.

As philosophical discipline Ontology (with a capital "o") is the study of the "*being qua being*" (Guarino 1995), i.e. the explanation of the reality via concepts, relations, and rules; the term ontology (with a lowercase "o") in the philosophical sense refers to a specific system of categories accounting for a specific idea of the world (e.g. Aristotle's ontology); in computer science an ontology refers to "An engineering artifact, constituted by a specific vocabulary used to describe a certain reality, plus a set of explicit assumptions regarding the intended meaning of the vocabulary words" (Guarino 1998, 4).

Ontologies enable to capture in an explicit and formal way the semantics of information sources. In a distributed environment such as the Web, resources are distributed and there is often the need to integrate different information in order to satisfy a user request. Semantic integration of information relies, currently, on ontologies integration: the process requires the identification of concepts similarity between ontologies and against user requests.

There are different ways of employing ontologies for information integration; Wache (2001) identifies three framework architectures:

- Single ontology approach: all information sources are related to one common global (top-level) ontology. The approach assumes that all information sources have nearly the same conceptualisation;
- Multiple ontologies approach: each information source has its own ontology. Since it cannot be assumed that these ontologies share the same conceptualisation, the lack of a common vocabulary makes the integration process difficult. In this case an inter-ontology mapping is required: this mapping tries to identify semantically correlated terms via semantic similarity measurement;
- Hybrid approach; hybrid approaches try to overcome the drawbacks of the two previous approaches. Each information source is described via its own ontology, but all source ontologies are built from the same global (top-level) ontology.

Semantic similarity measurements play a crucial role in this process, since they provide mechanisms for comparing concepts from different ontologies, and are, thus, the basis of semantic interoperability (a survey of different approaches to measuring similarity is provided in (Goldstone 2005)).

Since different ontologies may commit to different conceptualizations and may be implemented with different formalisms, inconsistency is inevitable. Ontology mismatches may be classified in two broad categories (Visser 1997):

- Conceptualisation mismatches: are inconsistencies between conceptualisations of a domain, which may differ in terms of ontological concepts or in the way these concepts are related (i.e. in a subsumption hierarchy);
- Explication mismatches: are related to the way the conceptualisation is specified, and occur when two ontologies have different definitions, but their terms, and concepts are the same.

The information integration process would be straightforward, at a certain degree, if all ontologies would be similar in terms of vocabulary, intended meaning, background assumptions, and logical formalism, but in a distributed environment this situation is hard, if not impossible, to achieve since different users have different preferences and assumptions tailored to their specific requirements in specific domains.

Notwithstanding the efforts to establish standards for ontology languages and basic top-level ontologies, there are still different approaches and heterogeneity between ontologies is inevitable (Krotzsch 2005).

Several authors (Goguen 2004, 2005, 2005a), (Krotzsch 2005), (Zimmermann 2006), are investigating the application of category theory to this issue. Goguen (2005) presents a theory of concepts (Unified Concept Theory – UCT) that integrates different approaches – Lattice of Theories (LOT), Formal Concept Analysis (FCA), Information Flow (IF), Conceptual Spaces, Conceptual Integration (Blending), Ontologies – while preserving their underlying conceptualizations. UCT approach to semantic integration uses category theory tools to unify all these approaches and generalises them to arbitrary logics based on the theory of institutions (Goguen 2004a).

Ontology-based approach to semantics is receiving also more foundational criticisms. Gärdenfors (2004) advocates, for example, that this approach is not very semantic; at the best it is ontological. Since it continues to build entirely on the symbolic methodology, it suffers from the symbol grounding problem; the question is how expressions in ontology languages may get any meaning beyond the formal language itself: ontologies are not grounded.

Another difficulty with ontologies is that they are decontextualised and decontextualise experience; instead <u>we claim that semantics reconciliation depends on contextual human sense-making</u>. Ontologies are forms of a priori agreements on a shared conceptualisation of a domain of interest, but meaning is an emergent feature arising context-dependently in action and acquired via participatory sense-making of socially coupled agents, rather than defined as symbolic rules (Di Paolo 2007), (De Jaegher 2007), (Flender 2008), (Dourish 2004).

Therefore, the use of ontologies is insufficient in a dynamic situation, where all interpretations may not be anticipated and on-the-fly integration may be needed (Ouksel 2003).

Several authors have proposed emergent semantics (Aberer 2004) as a solution; emergent semantics envisions a community of self-organising, autonomous agents interacting in a dynamic environment and negotiating meaning as required: this means that meaning emerges in context, but context itself is an emergent property, a feature of interaction of a community of practice (Dourish 2004).Emergent semantics is dynamic and self-referential, as a result of a self-organisation process; this requires some autonomous behaviour (cf. 2.2).

Collaborative tagging is a new paradigm of the web, where users are enabled to manage, share and browse collection of resources and to describe them with semantically meaningful freely chosen keywords (tags). These tags cannot even be considered as vocabularies, since there is no fixed set of tags nor explicit agreement on their use. Nevertheless, this set of unstructured, not explicitly coordinated tags evolves and leads to the emergence of a loose categorisation system (folksonomy) shared and used by a community of practice (Cattuto 2007).

Collaborative tagging falls within the scope of semiotic dynamics, i.e. the study of how a population of agents establish a shared semiotic system. Semiotic dynamics has been defined as "[...] the process whereby groups of people or artificial agents collectively invent and negotiate shared semiotic dynamics, which they use for communication or information integration" (Steels 2006, 32). Semiotic dynamics builds on different AI techniques, borrowing also ideas from the embodied hypothesis: the focus, however, is on social, collective, dynamic sense-making.

Computer simulations (Loula 2005), (Baronchelli 2006), (Puglisi 2008) have demonstrated that a population of embodied agents can self-organise a semiotic system.

## 4 Semantics and geographic information

Traditionally GIScience has relied on an objectivistic approach to knowledge creation (Schuurman 2006): this view assumes that GISystems represent the "real world" as it is independent of human cognition and perception of it (Schuurman 2002). The focus, therefore, has been ontological in nature more than epistemological; in determining geo-spatial ontologies, the question of "what exists" has gained much attention versus the question of "how what exists is identified and defined" (Agarwal 2005).

Starting from early 1990s, however, several researchers have focused their attention on the epistemological aspects of GIS, accounting for human cognition and perception. These authors borrow ideas from the experiential realism and the epistemological model introduced by (Lakoff 1987), arguing that cognition structures the perception and representation of reality: their work builds on image schemata, conceptual blending, conceptual spaces, and affordances.

Semantic issues have always been a key concern in GIS, since semantic interoperability plays a crucial role for the sharing and integration of geographic information (Harvey 1999). The use of ontologies is the most applied means to support semantic interoperability, and ontology has been recognized as a major research theme in GIScience (Mark 2000)

It is possible to individuate two main approaches to ontology in GIS:

- Philosophical approach: deals with top-level ontologies for the geographic domain, and takes an objectivistic view, i.e. reality as objectively existent independent of human cognition and perception. Works on this approach are, for example, (Mark 1999, 2001), (Smith 2001, 2004), (Frank 2001, 2007), (Kuhn 2003, 2005),(Galton 2003). Some of these authors highlight some issues, e.g. vagueness as well as cultural and subjective discrepancies that are difficult to solve: Mark (2003) has shown that people from different places and cultures use different categories for geographic features;
- Knowledge engineering approach: deals with ontologies as application-specific and purpose-driven engineering artifacts. Works on this approach are, for example, from (Kuhn 2001), (Camara 2000), (Hakimpour 2001), (Fonseca 2002), (Bernard 2003), (Klien 2004), (Raubal 2004), (Lutz 2006).

Frank (2001) suggests that an ontology for GIS should be built as a coordinated set of tiers of ontologies, allowing different ontological approaches to be integrated in a unified system constituted of the following tiers: (i) Tier 0 – human-independent reality; (ii) Tier 1 – observation of the physical world; (iii) Tier 2 – objects with properties; (iv) Tier 3 – social reality; (v) Tier 4 – subjective knowledge

Frank's five-tier architecture has a lot of commonalities with the four-universes paradigm (Gomes 1998) applied to geographic information by (Camara 2000a). Inspired by this last work, and based on a realistic view of the world, Fonseca (2002a), introduces a five-universes paradigm: (i) physical universe; (ii) cognitive universe; (iii) logical universe; (iv) representation universe; (v) implementation universe. Representing the reality involves the conceptualisation of elements of the physical world via a collective agreement of a community sharing common perceptions; concepts are defined within a community of experts and are organised in a logical framework (ontologies).

Halfway between a philosophical and an engineering approach, Kuhn (2003) proposes the Semantic Reference System as a framework for solving semantic interoperability problems. In analogy with spatial reference systems, semantic reference systems are composed of a semantic reference frame and a semantic datum: as the geometric component of geographic information refers to spatial reference systems, the thematic component refers to semantic reference systems. A semantic reference frame acts like a coordinate system, as a framework to which terms can refer to get meaning: this reference frame is a formally defined top-level ontology. As a datum in spatial reference systems anchors the geometry model to the real world, a semantic datum grounds the terms of semantic reference frame: Kuhn (2005) suggests using image schemas as grounding mechanism.

From a knowledge engineering point of view, ontologies have been applied for (i) geographic information discovery and retrieval (Bernard 2003), (Klien 2004), (Lutz 2006); (ii) geographic information integration (Sheth 1999), (Visser 2002), (Wache 2001), (Hakimpour 2001); (iii) GISystems (Fonseca 1999, 2002); (iv) modelling user activity (Camara 2000), (Kuhn 2001), (Raubal 2004), (Timpf 2002).

Application-specific, task-oriented, purpose-driven ontologies are aimed at information systems development: these ontologies emerge from requirements and contain knowledge limited to a specific area of application.

The main issue with engineering ontologies is grounding: according to the ontology hierarchy proposed in (Guarino 1998), domain and task ontologies are grounded in top-level ontologies. The question is how these top-level ontologies are grounded themselves: this infinite regress should end at some point to an ontology, but this is objectivism/reductionism and we are not in sympathy with this approach.

Borrowing ideas from embodied cognition and cognitive semantics, some authors propose image schemas as grounding mechanism and conceptual spaces as a new representation paradigm. Kuhn (2003), for example, proposes image schemas for grounding Semantic Reference Systems. Image schemas have been introduced in the geospatial domain by (Mark 1989), have received formal specifications in (Rodriguez 1997) and (Frank 1999), and have been applied, for example, to way finding in (Raubal 1998).

However, the concept of image schemas remain controversial and ambiguous, and disagreements exist in image schemas research between two broadly contrasting approaches: the first approach, located in the context of cognitive psychology and neurosciences, considers image schemas as expression of universal principles; the second approach, located in the context of anthropology and cognitive-cultural linguistics, has a more relativistic view of image schemas and emphasises that cognition is situated in socio-culturally determined context (Hampe 2005). Cross-linguistic and cross-cultural studies (Choi 1999) show that image schemas, while operating in many languages, are not universal, instead they are culturally situated (Correa 2005); on the other hand, the tendency to make universalistic statements is based on few languages, above all English (Zlatev 2007).

Some major drawbacks exist in the ontology approach to semantics in terms of dynamicity and context. Ontologies are a priori agreements on a shared conceptualisation of a domain of interest, a form of pre-given, decontextualised knowledge; this is problematic as long as we consider the temporal extent on the knowledge.

There is an empirical evidence of the fact that human manipulation of concepts is facilitated by considering them in relevant context (Barsalou 1993). Current approaches to context, also in the GI field (Rogriguez 1999), (Keßler 2007), are representational, i.e. they assume that context is stable, delimited information that can be known and encoded in just another information layer or another ontology in an information system.

Meaning, instead, is an emergent feature arising context-dependently in action and acquired via participatory sense-making of socially coupled agents, rather than as symbolic rules, and context itself is an emergent feature (Di Paolo 2007), (De Jaegher 2007), (Dourish 2004).

## 5 Toward a new formalism for geospatial semantics

We start from the assumption that meaning and context are dynamically emergent from activity and interaction, determined in the moment and in the doing (Dourish 2004): there is no pre-given knowledge and no fixed properties that can a priori determine what is relevant.

Current approaches to semantics, mainly ontology-based, are not able to satisfy and manage this assumption, and a new formalism is required.

The State-Context-Property (SCOP) formalism seems promising in this sense: a detailed description of the formalism can be found in (Aerts 2005, 2005a), (Gabora 2002, 2008), here a brief description is provided. SCOP is a formalism based on a generalisation of mathematics developed for quantum mechanics that provides a means for dealing with context, concept combination and similarity judgments, toward a contextual theory of concepts (Gabora 2008).

At conceptual level SCOP falls within the enactive approach to cognition: the main ideas behind it may be summarised as follows:

- Concepts as participatory (ecological view): concepts and categories are participating parts of the mind-world whole;
- Contextuality: context influences the meaning of concepts and needs to be given a place in the description of concepts;
- Concept combination: concept conjunctions exhibit emergent features that traditional theories (representational), are not able to "predict";
- Similarity: similarity judgments are context-dependent.

The description of a concept in SCOP consists of five elements:

- A set of states the concept may assume;
- A set of relevant contexts;
- A set of relevant properties;
- A function describing the applicability of certain properties in a specific state and context;
- A function describing the probability for one state to collapse into an other under the influence of a specific context.

For any concept we may have a number (infinite) of "possible" states, and each state is characterised by a unique typicality for instances and properties. A state of a concept that is not influenced by a specific context is said to be an *eigenstate*[1] for that context, otherwise it is said to be a *potential state* (superimposition state). A potential state may collapse to another state under the influence of a specific context: for example consider the concept **Tree** that under the context "**Desert island**" might collapse to state **Palm Tree**.

The properties of a concept are themselves potential: they are not definite except in a specific context. If a concept is in an *eigenstate* for a context, then the latter just detects what the concept is *in acto* (Gabora 2008), but if the concept is in a superposition state, then the context change this state: properties of the concepts may change under the influence of the context from actual to potential and vice versa. Therefore each state of a concept is an *eigenstate* or a superposition state: if it is an *eigenstate* the properties are actual, otherwise most of the properties are potential: a context has the "power" to change a superposition state in an *eigenstate*.

---

[1] An eigenstate is a state associated with definite properties

The various states of a concept can be described as a *Hilbert Space*, while the conjunction of two concepts can be modelled as an *entanglement*[2] through the tensor product of the two *Hilbert Spaces* describing those concepts.

Similarity judgments between two concepts are only possible if their respective properties are compatible, i.e. they refer to the same context (refer to Gabora 2002 for a detailed description of context-sensitive measure of conceptual distance).

## Future work

This work has dealt with semantics and how it is addressed in distributed information system and in the GI domain. Drawbacks of current practices, mainly based on ontologies, have been highlighted.

Ontology-based approach to semantics is problematic in terms of dynamicity and context, since ontologies, being forms of a priori agreements, are decontextualised.

Instead we claim for an enactive approach to cognition and semantics where meaning is an emergent feature arising context-dependently in action.

Since representational theories are unable to deal with these issues, a new formalism is required. The SCOP formalism is considered promising and has been briefly described.

Future work will deal with an in depth analysis of the formalism in order to investigate its applicability to GI, possibly toward practical applications.

---

[2] Entanglement means that two states are linked together and it is not possible to describe one state without mention of its counterpart